\documentclass[10pt,letterpaper]{article}

\usepackage{cogsci}
\usepackage{pslatex}
\usepackage{natbib}
\usepackage{comment}
\usepackage{url}

\usepackage{ccg-latex}
\usepackage{amsmath}
\DeclareMathOperator*{\argmax}{arg\,max}
\usepackage{graphicx}
\usepackage{floatrow}
\usepackage{subcaption}

\usepackage{tikz}
\usetikzlibrary{bayesnet}
\usepackage{pgfplots}
\usepackage{xcolor}
\definecolor{bblue}{HTML}{4F81BD}
\definecolor{rred}{HTML}{C0504D}
\definecolor{ggreen}{HTML}{9BBB59}
\definecolor{ppurple}{HTML}{9F4C7C}

\usepackage{todonotes}

\usepackage{hyperref}
\usepackage{cleveref}
\creflabelformat{equation}{#2#1#3}

\title{Word learning and the acquisition of syntactic--semantic overhypotheses}

\author{{\large \bf Jon Gauthier, Roger Levy, Joshua B. Tenenbaum} \\
  Department of Brain and Cognitive Sciences \\
  Massachusetts Institute of Technology \\
  \texttt{jon@gauthiers.net}, \texttt{\{rplevy,jbt\}@mit.edu}}

\begin{document}

\maketitle

\begin{abstract}
Children learning their first language face multiple problems of induction: how to learn the meanings of words, and how to build meaningful phrases from those words according to syntactic rules.
We consider how children might solve these problems efficiently by solving them {\em jointly}, via a computational model that learns the syntax and semantics of multi-word utterances in a grounded reference game. We select a well-studied empirical case in which children are aware of patterns linking the syntactic and semantic properties of words -- that the properties picked out by base nouns tend to be related to shape, while prenominal adjectives tend to refer to other properties such as color. We show that children applying such inductive biases are accurately reflecting the statistics of child-directed speech, and that inducing similar biases in our computational model captures children's behavior in a classic adjective learning experiment. Our model incorporating such biases also demonstrates a clear data efficiency in learning, relative to a baseline model that learns without forming syntax-sensitive overhypotheses of word meaning. Thus solving a more complex joint inference problem may make the full problem of language acquisition easier, not harder.

\end{abstract}

\section{Introduction}

Children face many distinct learning problems as they begin to understand and speak their first language. At the most fundamental level, they must infer the basic features of language: that words exist, and that they can be used to refer to entities in the world. At a higher level, they must work out what words actually mean, and how those words productively combine with one another syntactically to form phrases and sentences.

While every child seems to solve all of these difficult problems on her own, different lines of computational research have arisen to individually address some of these problems, each bringing their own respective toolset. Recent models of learning from noisy instances of perceptually grounded reference, for example, often leverage associationist or connectionist frameworks \citep{fazly2010probabilistic,chrupala2015learning,yu2004integration}, while studies of abstract rule learning about syntax and object category structure often rely on hierarchical Bayesian inference \citep{kemp2007learning,perfors2011learnability}. A productive line of work on the joint learning of lexical syntax and semantics has largely relied on one particular parsing formalism known as combinatory categorial grammar \citep{abend2017bootstrapping,steedman1996surface}.

This paper demonstrates how ideas from these separate subfields might be productively combined. We combine a model which incrementally learns the syntax and semantics of natural language with a process of structured probabilistic inference. This probabilistic inference allows the learner to induce linguistic overhypotheses --- abstract rules about the structure of language --- which the incremental learning model leverages to more efficiently generalize from new examples.

There is no shortage of such abstract rules about language which children observe. The shape bias \citep{smith2002object,kemp2007learning} is one of the strongest examples of an overhypothesis in language learning, according to which children reliably generalize labels for object categories based on the shape of their referents. Studies of child-directed speech have revealed that such an abstract rule might be a rational generalization from the data that children observe: many of the first nouns which children hear and produce refer to object categories which are defined by their shape \citep{samuelson1999early}.

We investigate a word learning effect which is related to the shape bias, but involves a more interesting interaction between syntax and semantics. \citet{smith1992count} presented children from ages 2;11 to 3;9 with novel objects, some of which shared the same shape with other objects and some of which shared the same color. In each trial, they labeled a novel object with either a noun frame (``this is a \emph{dax}'') or a prenominal adjective frame (``this is a \emph{dax} one''). The children were asked to then find all the other ``\emph{daxes}'' or ``\emph{dax} ones.'' Children took \emph{dax} to denote the shape of the novel object less often when it was labeled using the novel adjective versus the novel noun. This effect became stronger as \citeauthor{smith1992count} used objects and contexts which made the color of the referents more salient.\footnote{Other studies of adjective learning have found that, depending on the saliency of the relevant cue, children may also infer adjectives to denote properties relating to material and pattern \citep{taylor1988adjectives,mintz2002adjectives}. We do not account for these saliency cues in the model presented in this paper.}

The qualitative effect of interest is that children can appropriately interpret novel words in prenominal position to denote color properties while interpreting novel nouns to denote shape properties, modulating their inferences according to the cues of syntax. This effect provides us with a test case for presenting an enriched model of word learning, in which the learner induces probabilistic \emph{overhypotheses} relating abstract syntactic and semantic properties of the words it learns. Our combined model manages to capture the qualitative generalization behavior of children and demonstrates a more data-efficient language learning process.


\section{Corpus study}

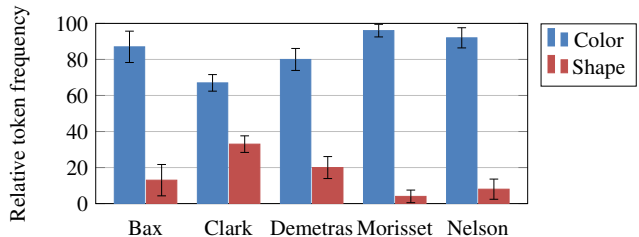
\begin{figure}
\resizebox{\linewidth}{!}{
\begin{tikzpicture}
    \begin{axis}[
        width  = \linewidth,
        height = 4.5cm,
        major x tick style = transparent,
        ybar=2*\pgflinewidth,
        bar width=14pt,
        ymajorgrids = true,
        ylabel = {Relative token frequency},
        symbolic x coords={
          Bax, 
          Clark,
          Demetras, 
          Morisset,
          Nelson
        },
        xtick = data,
        scaled y ticks = false,
        enlarge x limits=0.15,
        ymin=0, ymax=100,
        legend pos=outer north east
    ]
        \addplot[style={bblue,fill=bblue,mark=none}, error bars/.cd,
y dir=both,y explicit, error bar style={black}]
            coordinates {
              (Bax, 87) +- (8.7, 8.7)
              (Clark, 67) +- (4.6, 4.6)
              (Demetras, 80) +- (6.1, 6.1)
              (Morisset, 96) +- (3.5, 3.5)
              (Nelson, 92) +- (5.6, 5.6)
            };

        \addplot[style={rred,fill=rred,mark=none}, error bars/.cd,
y dir=both,y explicit, error bar style={black}]
             coordinates {
               (Bax, 13) +- (8.7, 8.7)
               (Clark, 33) +- (4.6, 4.6)
               (Demetras, 20) +- (6.1, 6.1)
               (Morisset, 4) +- (3.5, 3.5)
               (Nelson, 8) +- (5.6, 5.6)
             };

        \legend{Color,Shape}
    \end{axis}
\end{tikzpicture}
}
\caption{Relative frequencies of shape and color meanings denoted by prenominal adjectives in several corpora of child-directed speech. Error bars indicate standard error.}
\label{fig:childes-prenominals}
\end{figure}

We first examine whether the way children generalize the meanings of novel prenominal modifiers in these word learning experiments might be licensed given the language that they observe. If this were the case, we would expect to discover a reliable pattern among the properties denoted by prenominal modifiers. Concretely, given the experimental data discussed above, we should find that words in prenominal position are associated more often with color properties than with shape properties.

We evaluate this prediction in several corpora of child-directed speech, with English-speaking children ranging from age 1;9 to 3;11 \citep{bohannon1977children,clark1978awareness,demetras1986working,morisset1990environmental,nelson1989narratives}, with syntactic annotations automatically produced by \citet{sagae2007high}. For each corpus, we collected all instances of simple prenominal modification: phrases with the part-of-speech pattern \emph{\textless{}adjective\textgreater \textless{}noun\textgreater}, in which the adjective is in a direct dependency relation with the noun.
We manually classified each modifier we found as denoting a color property, shape property, or some other property.

\Cref{fig:childes-prenominals} shows the relative token frequency distribution of color and shape prenominal modifiers across multiple corpora of child-directed speech. We find the same pattern in each corpus: prenominal modifiers denote color properties far more often than they denote shape properties.\footnote{The corpus study also revealed other more common adjective meaning classes such as size terms and evaluative terms (``big,'' ``nice''). We focus on color and shape in this paper, which are the dimensions of meaning relevant to the generalization behavior of children described in \citet{smith1992count}. We are not aware of studies evaluating how children generalize with larger sets of salient dimensions, but are interested in testing how this broader set of corpus statistics drive children's meaning inferences in future work.}

This pattern suggests that children's preference to favor color interpretations when encountering prenominal modifiers, as reported in \citet{smith1992count}, could be the result of a bias learned from the statistical structure of child-directed speech. Children could exploit such statistical structure, that is, to form a belief linking the prenominal syntax of particular types of adjectives with certain types of property meanings.

\section{Reference game dataset}

\begin{figure}
\centering
\resizebox{0.5\linewidth}{!}{
\begin{tikzpicture}
\node [inner sep=0] (scene) at (0, 0) {\includegraphics[width=0.7\linewidth]{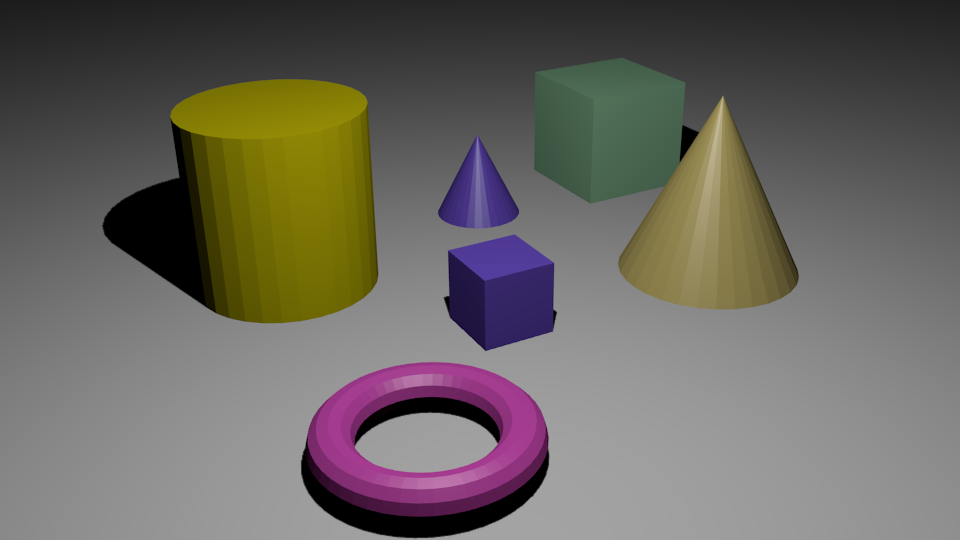}};
\draw [->, rred, line width=0.5mm] (-1.5, 1.9) -- (-0.1, 0.9);
\end{tikzpicture}
}
\caption{An example scene from our dataset. Given expressions such as ``the purple cone,'' the task of our language learner is to predict its referent by progressively learn the mappings between words and the properties they denote, based only past observations of utterances and their referents.}
\label{fig:scene}
\end{figure}

The findings of the previous section suggest that child-directed speech reliably associates syntactic properties (prenominal position) with semantic properties (color types). In the remainder of this paper, we present a computational model which automatically captures this statistical pattern as it incrementally learns words in a grounded reference game.

We design a reference game scenario in which a learner must pick out the intended referent of an utterance made in the context of some 3D scene. This type of reference game provides an interesting testbed for word learning models, since learners must generalize from instances of reference alone (without ever receiving explicit supervision on the meanings of the individual words that are used).

We construct a synthetic reference game dataset in which short utterances pick out objects in a 3D scene.\footnote{Because we do not have access to child-directed speech corpora with labeled ground-truth referents, we take the synthetic dataset as a first test of our model. The datasets available with more naturalistic language offer supervision only in the form of explicit meaning representations, which would not allow us to test the more challenging distantly-supervised learning setting proposed here.} \Cref{fig:scene} shows an example  scene in the dataset. In each of these trials, some utterance $\mathbf u = w_1, w_2, \dots, w_N$ is used to refer to a referent $R$, an object within the scene. The 3D objects in these scenes have several salient perceptual properties, including color and shape. The utterances in the dataset are of the form ``the $x$ $y$,'' in which $y$ is some noun and $x$ is a prenominal modifier.

We generate reference trials by first randomly sampling scenes with between 1 and 6 objects, each of which has a random color (out of 10 possible colors), shape (out of 10 possible shapes), material, and size. For each scene, we enumerate all possible referring expressions of the form ``the $x$ $y$'' which have a unique referent in the scene. Following the findings of the previous section, we enforce that the prenominal $x$ pick out some color property, and that the noun $y$ pick out some shape property.

The task of the learner in this dataset is to learn to predict the ground-truth referent $R$ in each reference trial, given just the utterance $\mathbf u$ and a representation of the objects in the scene. The learner must learn to make this prediction using supervision on the intended referents in each scene, without any access to the ground-truth meaning representations for each utterance.

\section{Incremental word learning model}
\label{sec:ccg}

We first formalize the problem of incrementally learning the meanings of novel words in the above dataset. At any given time $t$, we suppose that the learner has some lexicon $\Lambda_t$, linking wordforms to their inferred syntactic types and semantic interpretations. (The next paragraphs will define this lexicon in detail and show how it can be learned from data.) For each referring expression in our dataset, the learner observes that some utterance $\mathbf u_t = w_1, w_2, \dots, w_N$ is used to refer to a referent $R_t$, some object in the scene. The most interesting instances for our purposes are those in which the learner observes some novel word $w_i \not\in \Lambda_t$.

\newcommand\ltu{L_\mathbf{u}, T_\mathbf{u}}

\begin{figure}
\centering
\cgex{3}{the & blue & ball\\
\cglines{3}\\
\cgf{NP\fs NP} & \cgf{NP\fs NP} & \cgf{NP}\\
\lf{\iota(x)} & \lf{\lambda p. \lambda x.p(x) \land blue(x)} & \lf{\lambda x. sphere(x)}\\
&\cgline{2}{\cgfa}\\
&\cgres{2}{NP \lf{\lambda x. sphere(x) \land blue(x)}}\\  
\cgline{3}{\cgfa}\\
\cgres{3}{NP \lf{\iota(sphere(x) \land blue(x))}}
}\vfill
\caption{An example syntactic derivation of the phrase \emph{the blue ball} with the CCG formalism (read from top to bottom).}
\label{fig:ccg}
\end{figure}

We first introduce an incremental word learning model to address the problem formalized above. This model allows us to jointly predict syntactic and semantic analyses of sentences, and forms the base of our final model. It does not, however, have any capacity to implement overhypothesis-style beliefs. In the next section, we will augment this incremental word learner with a probabilistic model that supports such abstract rule learning.

The incremental learner implements the formalism of combinatory categorial grammar \citep[CCG;][]{steedman1996surface}. CCG is a bottom-up parsing method which jointly yields syntactic and semantic analyses of input sentences. For our purposes, a CCG grammar is a structure $G = (\Lambda, R)$ specifying a lexicon $\Lambda$ and a set of combinatory rules $R$. Each lexicon entry assigns a syntactic and semantic interpretation to a particular sequence of words. For example, the following lexicon is sufficient to yield an analysis of the phrase \emph{the blue ball}:
\begin{align*}
\text{the} &:= NP/NP &&: \lambda x.\iota(x) \\[-4pt]
\text{blue} &:= NP/NP &&: \lambda p. \lambda x. p(x) \land \text{blue}(x) \\[-4pt]
\text{ball} &:= NP &&: \lambda x. \text{sphere}(x)
\end{align*}

Each lexicon entry maps a token sequence to a syntactic type (e.g. $NP/NP$, $NP$) and a semantic expression represented here in the lambda calculus. The rule set $R$ specifies the legal ways in which entries from the lexicon may combine to produce constituent phrases. While CCGs support a substantial number of different rules, the only rule relevant to our particular task is that of \emph{forward application}. \Cref{fig:ccg} (read from top to bottom) shows how forward application is applied to derive an interpretation of the example sentence ``the blue ball.'' After first retrieving lexicon entries for each of the tokens in the sentence, we iteratively compose constituents of the form $NP/NP$ with arguments of type $NP$ appearing to the right. Whenever such syntactic composition occurs, we likewise compose the semantic expressions by function application.

The machinery presented so far allows us to derive bottom-up syntactic and semantic analyses of sentences. We call the final lambda calculus expression the \emph{logical form} of a sentence, and the particular sequence of rule applications the \emph{derivation} (analogous to a syntactic parse). For short, given an input sentence $\mathbf u$, let $L_\mathbf{u}$ denote the final logical form and $T_\mathbf{u}$ denote the derivation of the sentence.

\subsection{Probabilistic CCGs}

Because we wish to learn lexicon entries incrementally, we need to support a measure of uncertainty over the lexicon and the predictions of the CCG model. We associate a single parameter $\theta_j$ with each lexical entry $(w_j, s_j, m_j)$, which maps a particular wordform $w_j$ to a syntactic type $s_j$ and semantic interpretation $m_j$. With such a weighted lexicon, we can allow each wordform $w_j$ to have multiple possible interpretations, and learn to trade off these interpretations by training on language input.

The score of a particular sentence derivation $\langle \ltu \rangle$ simply depends on the particular lexical entries it draws on for each word.
These lexical scores are combined in a simple log-linear model, which can be efficiently and exactly computed via dynamic programming \citep{zettlemoyer2007online}:
\begin{equation}
P(\ltu \mid \mathbf u) \propto \exp\left(\sum_{(w_j,s_j,m_j,\theta_j)\in T_\mathbf u} \theta_j \right) \label{eqn:log-linear}
\end{equation}

\subsection{Validation-based lexical induction and learning}

While our CCG model is built to perform inferences over possible logical forms $L_\mathbf{u}$, its only source of supervision for any given instance is that an utterance $\mathbf u$ was used to refer to some referent $R$ within some grounded scene. In order to learn from such supervision, then, we rely on a deterministic \emph{validation function} $V(L_\mathbf{u})$ which returns the referent(s) denoted by $L_\mathbf{u}$ in the scene in which the sentence $\mathbf u$ was uttered.\footnote{A logical form inferred by the model may have zero to many referents depending on the objects visible to the listener and speaker.} Using this validation function, we implement a distantly supervised learning process after \citet{artzi2013weakly}, described below.

For each example utterance $\mathbf u$ and referent $R$, we first update the lexicon so that the sentence can be successfully parsed (see section ``\nameref{sec:ccg-induction}''). Using this augmented lexicon, we next retrieve all stored lexical entries for the words in $\mathbf u$.
Given these sets of weighted lexical entries, we perform the inference $P(\ltu \mid \mathbf u)$ using the CYK algorithm. We finally use the ground-truth referent $R$ in order to increase the relative probability of parses $L_\mathbf{u}$ where $V(L_\mathbf{u}) = R$ (see section ``\nameref{sec:ccg-learning}'').

\subsubsection{Lexical induction}
\label{sec:ccg-induction}

Suppose we encounter sentence $\mathbf u$ used to refer to some referent $R$ which the model cannot parse using the lexicon $\Lambda$. This may be because the sentence contains a novel word, or because none of the existing lexical entries can combine to produce the desired logical form. In either case, our job is to infer candidate syntactic types and meaning interpretations for the words in question which help us map the sentence $\mathbf u$ to the referent $R$. Let $w_i$ denote the word which requires novel candidate interpretations. We enumerate all possible syntactic and semantic interpretations of the word $w_i$, subject to two critical constraints:
\begin{enumerate}
\item The syntactic type of $w_i$ must yield a valid parse when combined with the other words of the sentence.
\item The semantic interpretation of $w_i$ must yield a logical form $L_\mathbf{u}$ when composed with the other words of the sentence such that $V(L_\mathbf{u}) = R$.
\end{enumerate}

For example, suppose that we observed the sentence shown in \Cref{fig:ccg} used to refer to a blue ball, but did not have the words ``blue'' or ``ball'' in our lexicon. Given the single available syntactic rule of forward application in our model, we would immediately be able to conclude that the syntactic type of ``blue'' was $NP/NP$ --- a prenominal modifier --- and that the syntactic type of ``ball'' was $NP$. The semantics are less constrained, however: we might plausibly infer that ``blue'' denotes a property \emph{blue} and ``ball'' denotes a property \emph{sphere}, but we could also infer that ``blue'' means \emph{sphere} and ``ball'' means \emph{blue}. Both interpretations would combine to yield a semantic analysis which predicts the correct referent.

We temporarily insert all of these possible interpretations into the lexicon with associated zero weights.
Using this augmented lexicon, we can find the logical form and derivation $\ltu$ which maximize \Cref{eqn:log-linear}. Given this maximal parse, we retrieve the interpretations of the novel word(s) in question and permanently add them to the lexicon.

\subsubsection{Learning}
\label{sec:ccg-learning}

The parameters of the lexicon $\Lambda$ are optimized with a stochastic online learning algorithm, as described in \citet{artzi2013weakly}. 
We perform a perceptron update on the lexicon weights in order to maximally separate the computed scores of parses which yield the correct referent $R$ from those which yield the incorrect referent.\footnote{See \citet{artzi2013weakly} for algorithm details.}

\section{Overhypothesis model}

The model presented so far provides a framework for incrementally learning a lexicon and predicting syntactic and semantic analyses of sentences. In order to capture the phenomenon of interest in this paper, we need to describe how this process of word meaning inference can be guided by abstract beliefs about the structure of that induced lexicon.

This section proposes an augmented meaning inference process for novel words within the CCG framework. It leverages the existing knowledge represented in the CCG lexicon to incrementally track beliefs about the abstract relationships between certain syntactic patterns and semantic features. This belief is precisely the overhypothesis which we expect to link prenominal syntax to particular types of property semantics.

\subsection{Lexical induction with priors}

\begin{figure}
\floatbox[{\capbeside\thisfloatsetup{capbesideposition={left,top},capbesidewidth=5.4cm}}]{figure}[\FBwidth]
{\caption{The probabilistic model used to predict the semantic properties of a word given its syntactic properties and surface form. Shaded circles are observed variables and empty circles are latent. See main text for details.}\label{fig:plate}}
{\begin{tikzpicture}
  \node[latent] (ftype) {$t$};
  \node[latent, right=of ftype] (fval) {$v$};

  \node[obs, below=of ftype] (syntax) {$s$};
  \node[obs, below=of fval] (word) {$w$};

  \edge {ftype} {syntax};
  \edge {ftype} {fval};
  \edge {fval} {word};

\end{tikzpicture}}
\end{figure}
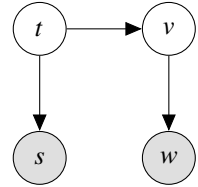

Note that the CCG model presented so far treats all novel candidate interpretations of a word equally: it simply attempts to produce derivations with all syntactically legal interpretations of the word, and picks the top-scoring derivation which predicts the correct referent in a scene.

We modify this lexical induction process by using a probabilistic model to seed initial weights for these candidate interpretations. In our case, each candidate interpretation of a word $w$ is of syntactic type $s \in \{NP,\, NP/NP\}$, and refers to some object property value (e.g. \emph{sphere} or \emph{blue}). Each object property has some abstract type $t$ (e.g. shape or color). Our model assigns initial weights to these candidate interpretations by predicting a distribution over the possible property types and particular property values of a word $w$ with a candidate syntactic interpretation $s$, $P(t, v \mid s, w)$.

\Cref{fig:plate} presents a diagram of the probabilistic model powering this prediction. Here $P(t)$ is a uniform prior over property types and $P(v \mid t)$ is a deterministic map, associating particular property values with particular property types.
The remaining distributions --- $P(s \mid t)$, which specifies how particular property types map to particular syntactic types, and $P(w \mid v)$, which specifies how particular property values map to particular words, can be calculated simply by inspection of the weighted CCG lexicon. We use the lexicon weights to build Dirichlet priors on these distributions as follows:
\begin{align}
\alpha_{s \mid t} &\propto \exp\left(\frac{1}{\tau} \sum_{(w_i, s_i, m_i, \theta_i) \in \Lambda} \mathbf 1\{\text{PT}(m_i) = t \land s = s_i\} \theta_i \right) \label{eqn:alphas}\\
\alpha_{w \mid v} &\propto \exp\left(\frac{1}{\tau} \sum_{(w_i, s_i, m_i, \theta_i) \in \Lambda} \mathbf 1\{\text{PV}(m_i) = v \land w = w_i\} \theta_i \right) \label{eqn:alphaw}
\end{align}

Here $\text{PT}(m_i)$ represents the property type associated with a word meaning $m_i$, and $\text{PV}(m_i)$ likewise represents the property value. The summand in \Cref{eqn:alphas} simply aggregates the weights of all lexicon entries whose meaning refers to the attribute type $t$ and whose syntactic type is $s$. Likewise, the summand in \Cref{eqn:alphaw} aggregates the weights of all lexicon entries which refer to an attribute value $v$ and have surface form $w$. These counts are then scaled according to a temperature parameter $\tau$, and 
 are used to predict conditional distributions with mass parameters $\rho_s, \rho_w$:
\begin{equation}
P(s \mid t) \sim \text{Dirichlet}(\rho_s \alpha_{s \mid t});
P(w \mid v) \sim \text{Dirichlet}(\rho_w \alpha_{w \mid v})
\end{equation}

The mass parameters $\rho_s, \rho_w$ determine how strongly these induced counts influence the distributions $P(s \mid t)$ and $P(w \mid v)$. As each $\rho$ term increases, the learner's prior beliefs are more likely to influence their inferences about novel words.

\subsection{The integrated learning process}

These predicted weights $P(t, v \mid s, w)$ are used to assign initial scores to candidate interpretations of novel words. This allows the probabilistic model --- which tracks beliefs over time about the relationship between syntactic properties and semantic properties --- to influence how novel words are interpreted by the learner. Concretely, the integrated learning process works as follows:
\begin{enumerate}
\item Given an utterance $\mathbf u$ with referent $R$, perform validation-based lexical induction to derive a set of candidate syntaxes and meanings for any words which are missing sufficient lexical entries to map $\mathbf u \to R$. Each candidate meaning contains some property with type $t$ and value $v$.
\item Use the inferred distribution $P(t, v \mid s, w)$ to assign weights to these candidate meanings.
\item Select the candidate meaning which yields a parse with the maximal weight $P(\ltu \mid \mathbf u)$ as in \Cref{eqn:log-linear}, and augment the lexicon accordingly. Retain the weights for the winning lexical entries as given by $P(t, v \mid s, w)$.
\item Update the weights of the augmented lexicon, favoring derivations which yield the correct referent $R$.
\end{enumerate}

The probabilistic model $P(t, v \mid s, w)$ is thus fully integrated into the existing CCG learning mechanism. Its inference is driven by prior distributions derived from the lexicon, and the results of its predictions provide the initial weights for each new lexical entry. Because this probabilistic model tracks an abstract property of the lexicon --- namely, the correspondence between syntactic type and semantic features --- the CCG learner can begin to exploit any such correspondence in the data.

\section{Results}

We evaluate the proposed overhypothesis model on our synthetic dataset of referring expressions. The CCG parsing and learning algorithm is implemented using a fork of the Cornell Semantic Parsing Framework \citep{artzi16spf}, and we implement the overhypothesis inferences using WebPPL \citep{webppl}.

We compare the overhypothesis model to a standard CCG learner, referred to in this section simply as the \emph{base model}. This model simply randomly initializes the weights of the candidate meanings it entertains, rather than using the probabilistic inference described in the previous section to power meaning inferences. Without such integrated probabilistic inference, the base model has no capacity to exploit the structure of the data relating syntactic and semantic properties.

\begin{figure*}
\begin{subfigure}{0.49\linewidth}
\includegraphics[width=\linewidth]{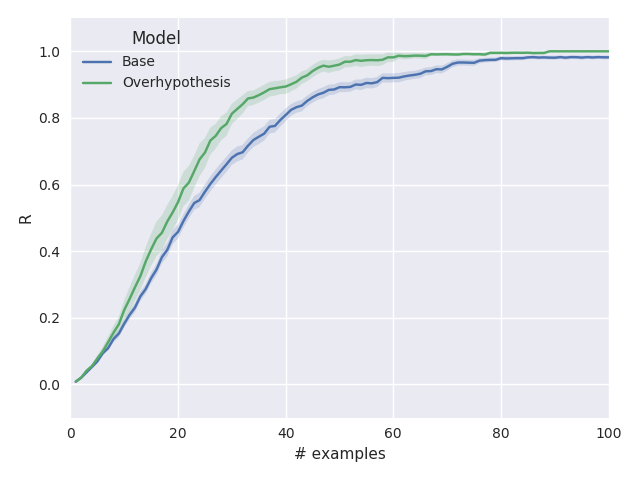}
\caption{Online accuracy curves (see \Cref{eqn:accuracy}) estimated on a fixed test set after each training example. Shaded region represents bootstrap estimate of 95\% CI.}
\label{fig:online-results}
\end{subfigure}\hfill%
\begin{subfigure}{0.49\linewidth}
\includegraphics[width=\linewidth]{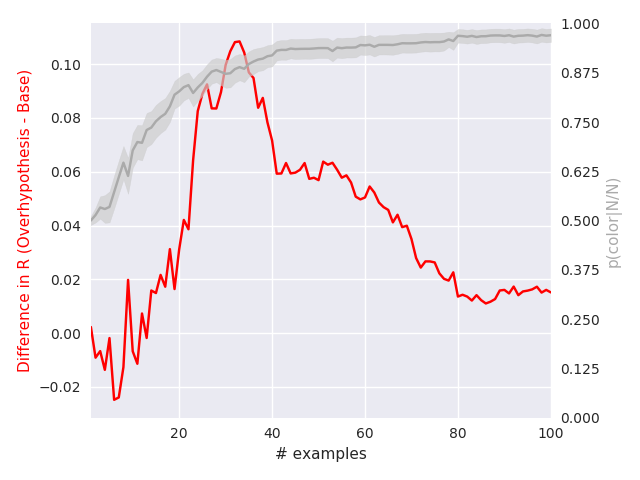}
\caption{Syntactic--semantic overhypotheses support rapid learning. The red line (left axis) plots the mean difference in $R$ between runs of the overhypothesis model and runs of the base model as learning progresses. The gray line (right axis) tracks the average strength of the induced overhypothesis $p(t = \text{color} \mid s = N/N)$ in the overhypothesis model. Shaded region represents bootstrapped 95\% CI.
}
\label{fig:overhypothesis}
\end{subfigure}
\caption{Results from the computational simulation of syntactic--semantic overhypothesis learning.}
\end{figure*}

\subsection{Predictive performance}

We first evaluate how well the two models perform in predicting the referents of utterances in our synthetic dataset. We evaluate each model after every learning trial on a fixed dataset, computing the following online accuracy metric:
\begin{equation}
\begin{aligned}
R &= \frac 1 M \sum_{i=1}^M \mathbf 1\{V(\argmax_{L_\mathbf{u}} P(\ltu \mid \mathbf u_i)) = R_i\} \\
  &= (\text{\# correct referent predictions}) / (\text{\# examples})
\end{aligned}
\label{eqn:accuracy}
\end{equation}

\Cref{fig:online-results} shows this online metric computed across many random restarts of both the overhypothesis model and the base model. In each random restart, the sequence of scene and utterance pairs in the dataset is randomly reshuffled. We run many instances of each model on these shuffled sequences, and use the results from different runs of the sample model to estimate 95\% confidence intervals on the metric $R$.

The gap between the learning curves in \Cref{fig:online-results} shows that the overhypothesis model benefits from an increased data-efficiency over the base model, and that this data-efficiency increases over time.

~

\Cref{fig:overhypothesis} illustrates the source of this data efficiency. The red curve (left axis) plots the performance gap: the average difference in the metric $R$ between the overhypothesis model and the base model at each timestep (the gap between the two curves in \Cref{fig:online-results}). The gray curve (right axis) tracks the average belief in the relevant overhypothesis for this model, as represented across many different learning runs. We compute this belief by recording the model's predictions of the quantity $p(t = \text{color} \mid s = N/N)$ after each learning instance. (This quantity depends directly on the model weights described in \Cref{eqn:alphas}.)

At the beginning of the learning process, the gray line shows that $p(t = \text{color} \mid s = N/N) = 0.5$ --- the average model has no belief in the overhypothesis. After just a few examples, the overhypothesis belief rapidly strengthens. By 10 examples, the perfomance gap begins to climb along with the overhypothesis belief curve. The performance gap reaches a maximum difference in $R$ of about 11\%, after which the base model slowly begins to approach the performance of the overhypothesis model.

The trend of the gray line also confirms the model's qualitative fit: by inducing the overhypothesis, our model is increasingly likely to infer that novel prenominal modifiers denote some instance of a color property.

\section{Discussion}

This paper connects to a large body of recent work on computational models of word learning. Most relevant is the work of \citet{abend2017bootstrapping} and \citet{sadeghi2018early}, who present models which jointly learn both word meanings and syntactic properties of language. Their models demonstrate how syntactic knowledge can be leveraged to more quickly infer the meanings of words.
Our contribution is complementary: we introduce more complex structured probabilistic models supporting a class of overhypothesis linking syntactic properties of words to certain aspects of their meaning.



Our model is an initial demonstration of how overhypotheses are a clear adaptive feature for a language learner, and how they can be tractably modeled by linking distinct computational tools. While there is naturally more to children's adjective learning than the phenomenon described in this paper, we believe the results support a general conclusion about the structure of the language learning problem. By explicitly tracking abstract relations between the syntactic and semantic properties of words at the level of the lexicon, a language learner can become more efficient at acquiring word meanings over time. Our computational model demonstrates how such abstract relations can be induced from very little data, and learned in an efficient and incremental fashion.

\bibliographystyle{apalike}

{\small
\bibliography{CogSci_Template}}

\end{document}